# RHealthTwin: Towards Responsible and Multimodal Digital Twins for Personalized Well-being

Rahatara Ferdousi, *Member, IEEE*, M. Anwar Hossain *Senior Member, IEEE*

*Abstract* — The rise of large language models (LLMs) has created new possibilities for digital twins in healthcare. However, the deployment of such systems in consumer health contexts raises significant concerns related to hallucination, bias, lack of transparency, and ethical misuse. In response to recommendations from health authorities such as the World Health Organization (WHO), we propose Responsible Health Twin (RHealthTwin), a principled framework for building and governing AI-powered digital twins for well-being assistance. RHealthTwin processes multimodal inputs that guide a health-focused LLM to produce safe, relevant, and explainable responses. At the core of RHealthTwin is the Responsible Prompt Engine (RPE), which addresses the limitations of traditional LLM configuration using four controlled modules. Unlike conventional LLM interfaces where users input unstructured prompts and instructions, increasing the risk of hallucination, RPE dynamically extracts predefined slots to structure these inputs. As a result, generated responses are context aware, personalized, fair, reliable, and explainable for well-being assistance. The framework further adapts over time through a feedback loop that updates the prompt structure based on user satisfaction. We evaluate RHealthTwin across four consumer health domains including mental support, symptom triage, nutrition planning, and activity coaching. RPE achieves state-of-the-art results with BLEU = 0.41, ROUGE-L = 0.63, and BERTScore = 0.89 on benchmark datasets. Also, we achieve over 90% in ethical compliance and instruction-following metrics using LLM-as-judge evaluation, outperforming baseline strategies. We envision RHealthTwin as a forward-looking foundation for responsible LLM-based applications in health and well-being.

*Index Terms* — Bias mitigation, Consumer Health, Digital twins, Ethical artificial intelligence, Explainable artificial intelligence, Generative AI, Healthcare informatics, Large language models, Multimodal systems, Personalized healthcare, Responsible AI, WHO guidelines You can explore the additional materials and prototype at:

https://github.com/turna1/ResponsibleHealthTwin-RHT-
https://huggingface.co/spaces/Rahatara/WellebeingDT

## I. Introduction

Large Language Models (LLMs) have begun to transform healthcare and well-being applications by serving as interactive medical assistants, health education tools, and patient engagement platforms. For instance, specialized LLMs like Google's Med-PaLM [1] have been fine-tuned to answer medical questions with high accuracy on U.S. Medical Licensing Exam (USMLE) questions. General-purpose LLMs (e.g. ChatGPT) are likewise being explored for clinical decision support and patient Q&A, showing the potential of conversational AI to improve patient engagement and health literacy when used carefully [2] [3].

In general Digital Twins (DTs) represent virtual replica of an entity or process in real world. Recent research even uses LLMs to generate DT of patients for clinical simulations for example, the TWIN-GPT system leverages ChatGPT's vast medical knowledge to create unique, personalized digital patient profiles that can aid virtual clinical trials [4]. These advances underscore a broad scientific precedent that LLMs can encapsulate healthcare knowledge and reasoning ability at or near the level of human experts opening new frontiers in personalized healthcare AI [5]. While this existing DT approaches focus on synthetic patient record generation [4], clinical process simulation [6], disease diagnosis [7] , these implementations often neglect user engagement, ethical safeguards, and adaptability for everyday health tasks [8]. There remains a clear gap in creating context-aware, explainable, and ethically compliant DTs for non-clinical, consumer-oriented use cases.

Also, LLMs often produce inaccurate or hallucinated content, show sensitivity to prompt phrasing, and may inherit societal or demographic biases from their training data. These risks are particularly acute in health settings, where misleading outputs can directly affect human well-being. Furthermore, studies have shown that health professionals and patients may exhibit automation bias, deferring critical judgment to AI systems. This increases the likelihood of over-reliance on flawed recommendations [9]. The World Health Organization (WHO) has accordingly issued a cautionary stance on using LLMs for healthcare, outlining six guiding ethical principles: autonomy, well-being, explainability, accountability, equity, and sustainability [10].

This ethical imperative motivates us to design a framework align with WHO's ethics principle for responsible practice to use multimodal LLMs in well-being interventions. We propose, a novel framework that combines LLM-based generation with ethical prompt governance and feedback-driven interaction. At the core of the system is a Responsible Prompt Engine that integrates: 1) Context-Aware Task Personalization Module to adapt queries based on individual goals and temporal health data; 2) Adaptive System Behavior Management Module to dynamically configure the model's role, tone, and accountability mechanisms; 3)Filter Constraints Module to ensure bias mitigation, safety, and fairness; 4) Justification and Grounding Module to generate evidence-backed, explainable outputs using internal memory and retrieval from domain knowledge bases. A bidirectional feedback loop between the user (in real world) and the system (digital twin) further enables refinement of recommendations over time, creating a personalized, evolving companion. The generated DT from this framework is an actionable well-being assistant for daily health management, lifestyle planning, and self-guided health education.

To rigorously assess the effectiveness of RHealthTwin, we conducted a structured evaluation across four publicly available well-being datasets: MentalChat16k [11][1], MTS-Dialog [12][2], NutriBench(Version 2) [13][3], and SensorQA [14][4]. Moreover based on these datasets, we synthetically generated over 4,000 test prompts simulating both patient and provider perspectives to evaluate responsible behavior in real-world health interactions. We adopted and formalized key metrics including Factuality Score (FS), Contextual Appropriateness Score (CAS), Instructional Compliance Score (ICS), and WHO-aligned Responsibility Rubric (WRR). Leveraging GPT-4 as an evaluator function enabled scalable, consistent judgment of response quality and ethical compliance across prompt strategies. Our evaluation confirms the impact of the Responsible Prompt Engine in producing structured, trustworthy, and context-aware outputs essential for personalized well-being digital twins.

We evaluated the RHealthTwin on four health datasets: MentalChat16k, MTS-Dialog v3, NutriBench v2, and SensorQA. These cover mental health, lifestyle, nutrition, and clinical QA. We generated over 4,000 question-based prompts from patient and provider perspectives on these datasets. We used standard metrics: BERT Score, BLEU, and ROUGE-L. We also proposed four key metrics: Factuality Score (FS), Contextual Appropriateness Score (CAS), Instructional Compliance Score (ICS), and WHO-aligned Responsibility Rubric (WRR) based on trending LLM as evaluator approach.

On datasets with ground-truth references (e.g., MentalChat16K and MTS-Dialog), RPE also produced the highest reference-based scores, including BLEU (0.41), ROUGE-L (0.63), and BERTScore (0.89). This indicates that RPE helps to align the generated output of RHT (e.g., decision making, risk identification) closely with human-annotated responses both lexically and semantically. RHealthTwin outperformed zero-shot, few-shot, and instruction-tuned baselines. Through overall experimental evaluation, we found RHealthTwin enables reliable, explainable, and ethically aligned response generation across mental health, nutrition, lifestyle, and clinical QA domains. To assess responsible behavior, we used ICS and WRR. RPE consistently outperformed all baselines, achieving ICS > 0.94 and WRR > 0.92 across datasets. In MTS-Dialog, for instance, RPE scored ICS = 0.947 and WRR = 0.928, compared to ICS = 0.816 and WRR = 0.775 from instruction-tuned prompts. These gains held across both patient and provider roles, reflecting RPE's ability to enforce tone, safety, and ethical alignment. Unlike traditional few-shot prompting, which showed inconsistent tone and safety control (ICS often < 0.85), RPE maintained ethical fidelity under all evaluation conditions.

**Our key contributions are as follows:**

1) RHealthTwin as the first (to the best of our knowledge) LLM-driven digital twin framework for responsible consumer health and well-being applications.
2) A slot-based responsible prompt engine that dynamically extracts context, constraints, and justification from user input, enabling personalized yet ethically bounded LLM responses.
3) Algorithms for transforming unstructured user query input into structured prompt and system instruction to facilitate adaptive AI inference.
4) Synthetic test prompts from four benchmark healthcare datasets, covering domains such as mental health, clinical question-resolution, nutrition, and lifestyle monitoring, to allow a realistic and role-specific evaluation of the framework.

The remainder of this paper is organized as follows. Section II, presents a literature review, Section III, describes the system methodology and component design. Section IV, outlines experimental setup, datasets, and evaluation. Section V, discusses findings in the context of responsible AI design. Finally, Section VI, concludes with future directions.

## II. LITERATURE REVIEW

In this section, we review recent advancements in large language models (LLMs) for healthcare applications. We then summarize key principles for responsible AI in the healthcare domain. Finally, we highlight gaps and remaining challenges in the existing literature that guide the design of the RHealthTwin framework.

### A. LLM and Digital Twin Integration in Healthcare

DTs in healthcare are virtual models of a patient or system that fuse real-time data and simulations to personalize treatment [7] [15]. As tabulated in Table I, existing study on healthcare DTs highlights their potential to predict, prevent, and manage disease by continuously integrating multimodal patient data. Recent research has begun to leverage LLMs as the core engine of such twins, enabling natural-language interaction and reasoning.

For example: TWIN-GPT [4] uses a fine-tuned ChatGPT model to create personalized digital patient twins for clinical trial simulation. Given sparse trial outcome data, TWIN-GPT encodes patient EHR features in LLM prompts to generate synthetic, patient-specific histories. This LLM-generated twin preserves individual characteristics and improved trial outcome prediction compared to prior models. Another study proposes DT-GPT [6] to fine-tune a biomedical LLM (BioMistral) on longitudinal EHR data to forecast patients' lab trajectories. Inputs (demographics, vitals, labs) are templated into text prompts; the LLM then predicts future labs in a zero-shot chatbot mode. DT-GPT achieved state-of-the-art forecasting accuracy on cancer and ICU datasets, demonstrating that an LLM-based twin can anticipate patient health states. HDTwin [7] integrates heterogeneous cognitive health data into a "human digital twin". It converts diverse inputs (demographics, sensor-derived behavior markers, speech transcripts, EMA responses) into text and fetches relevant medical literature via a retrieval pipeline.

---

[1] MentalChat16K dataset is released at https://github.com/ChiaPatricia/MentalChat16K_Main.
[2] MTS-Dialog v3 dataset can be accessed from https://github.com/abachaa/MTS-Dialog.
[3] NutriBench dataset is available on Hugging Face at https://huggingface.co/datasets/dongx1997/NutriBench.
[4] SensorQA dataset is publicly available at https://github.com/benjamin-reichman/SensorQA.

TABLE I
COMPARATIVE ANALYSIS OF LLM-DRIVEN DIGITAL TWIN FRAMEWORKS IN HEALTHCARE

| Aspect | TWIN-GPT | DT-GPT | HealthLLM | PsyDT | RHealthTwin |
|---|---|---|---|---|---|
| Application | Clinical trial simulation | Disease trajectory forecasting | Wearable-based disease prediction | Mental health journaling and therapy recommendation | Consumer well-being, daily health planning, journaling, scenario support |
| Data Sources | Structured EHR (longitudinal) | Multimodal (EHR + vitals) | Reports + wearable metrics | Text-based patient logs + affect data | Sensor data, text/image input, EHRs |
| LLM Role | EHR synthesis for counterfactual evaluation | Time-aware encoder + LLM forecasting | LLM for feature prediction via RAG | LLM for therapeutic dialogue and risk stratification | Instruction-tuned LLM with task context, justification, and explanation output |
| Personalization | Per-patient simulation | Temporal profile encoding | Risk scores from wearable logs | User-centric adaptation | User-centered dynamic system role generation |
| Ethical Focus | Limited (data privacy) | Limited (model fairness) | Partial (calibration) | Awareness of stigma | Embedded WHO-compliant modules (fairness, transparent, safety, accountability, human well-being) |
| Trust Mechanisms | None | Simple scoring metrics | Clinician scoring of RAG answers | System Instruction | Prompt modules (instruction, filter, explanation); with constraints |
| Hallucination Handling | None | Regularization only | RAG with clinical references | Not addressed | Grounded filtering and fallback |
| Feedback Loop | None | No real-time interaction | Static mapping | Manual journaling input, not bidirectional | Real world–digital twin interaction for updates, correction, and learning |
| Multimodal Reasoning | Limited | Structured + tabular + time-series | Sensor-focused | Single modality (text only) | Fully multimodal: images, sensors, text, logs |
| Explainability | Not included | Time-step loss only | Black-box with minor evaluation | Not transparent | Justification module with source-linked references |
| Regulatory Alignment | Not addressed | No specific compliance framework | No regulatory auditability | Mental health sensitivity only | Fully aligned with WHO ethics (autonomy, explainability, equity, etc.) |

A LangChain-GPT-3.5 system then reasons over this multimodal context to predict cognitive diagnoses and explain its reasoning. HDTwin's prototype outperformed ensemble baselines, yielding higher accuracy (~0.81) in detecting mild cognitive impairment, while providing interactive, explainable chat about the patient formative. A schematic of HDTwin's architecture shows how personal markers, statistics, and scientific papers are retrieved and aggregated into a prompt for the LLM. PsyDT (Psychological Digital Twin) uses GPT-4 to build a conversational twin of a specific counselor. It synthesizes thousands of multi-turn therapy dialogues by first modeling each counselor's personal linguistic style and treatment approach, then generating client–counselor exchanges that match real cases. The resulting LLM (PsyDTLLM) was fine-tuned on these synthetic chats and can emulate a therapist's style in novel sessions. This work exemplifies using an LLM to create a mental-health digital twin, facilitating personalized counseling simulations.

Existing LLM-based digital twins primarily focus on clinical applications, such as: DT-GPT's EHR-based trajectory forecasting for ICU/lung cancer patients or HDTwin's cognitive health diagnostics. These systems excel in technical accuracy, but face critical gaps in ethical governance, consumer accessibility, and dynamic personalization. In contrast, our proposed RHealthTwin framework addresses these gaps by integrating (1) a responsible prompt engine to enforce transparency and bias mitigation (2) multimodal grounding combining wearables, user queries, and health knowledge, and (3) feedback-driven adaptation to refine outputs iteratively.

### B. Trust in Healthcare LLMs

Building user trust in LLM-based health tools requires rigorous alignment of model behavior to clinical needs. Common strategies include instruction tuning, prompt tuning, and few-shot prompting to better align LLM responses with desired tasks. Instruction tuning involves fine-tuning an LLM on a large dataset of "instruction–response" pairs so it learns to follow healthcare-specific commands. For instance, Wu et al.

TABLE II
OPERATIONALIZATION OF WHO RESPONSIBLE AI PRINCIPLES IN RESPONSIBLEHEALTHTWIN FRAMEWORK.

| WHO Principle (2021) | Guidance | LLM-Related Risk | Design Safeguard | Requirements for RHealthTwin |
|---|---|---|---|---|
| Autonomy | Humans must remain in control; valid informed consent, privacy, and data protection are required | Over-reliance on AI; opaque use of patient data; consent bypass | Explicit consent mechanisms, user control over data use, opt-in actions | Consent-aware prompt governance; user feedback integrated into system memory |
| Well-being & Safety | AI must be safe, accurate, and evaluated under real-world use conditions for each task | Hallucinated advice, incorrect medical information, unsafe suggestions | Accuracy auditing, scope definition, context-aware completions | Responsible Prompt engine for controlled suggestion space, wellness-oriented task curation |
| Transparency & Explainability | Designs must disclose sufficient technical information for meaningful review and understanding | Black-box responses; no clarity on decision rationale | Reasoning, output citation, rationale generation | Justification & Grounding module |
| Accountability | Stakeholders must be liable; systems must include redress pathways and auditability | No logs of decision processes; unclear responsibility if errors occur | Trace logging, role specification, human-in-the-loop resolution | Role-based system behavior module; loggable system messages and instruction tuning traceability |
| Equity & Inclusiveness | AI must not exacerbate health disparities; should serve all populations fairly | Biased outputs from skewed training data; marginalization of underrepresented users | Demographic evaluation, re-balancing with synthetic data, multilingual access | Synthetic user-centric few-shot example generation for diverse profiles; demographic fairness score embedded in filter prompt module |
| Sustainability & Responsiveness | AI must be maintained, retrained, and evaluated to ensure long-term alignment; impacts on labor must be mitigated | Static models, energy waste, exclusion of workforce in design and use | Modular architecture, energy-efficient deployment, continuous evaluation | Feedback loop and iterative prompt retraining; mobile-deployable AI with real-world feedback. However, the impact of workforce remains challenging to address at this early stage of adoption. |

compiled MIMIC-Instr, a dataset of over 400,000 open-ended instructions derived from EHR data, and used it to tune an LLM (Llemr) for clinical tasks. This approach produced a conversational EHR assistant whose performance matched or exceeded specialized baselines [16].

Prompt tuning (training soft or prefix prompts) is another lightweight adaptation: in medical QA, carefully engineered prompts or small prompt parameters can steer an LLM's outputs toward accuracy [17]. For example, recent work showed that open-source LLMs tuned via prompt engineering can sometimes outperform full-model fine-tuning on medical QA benchmarks, emphasizing the value of domain-specific prompting techniques [18].

Few-shot prompting (providing a handful of exemplars in the prompt) is also commonly used; by demonstrating the format of desired answers, it helps guide the model without full retraining. These few-shot strategies leverage the LLM's existing knowledge while requiring minimal new data [19].

Beyond technical tuning, human-in-the-loop oversight and guidelines-based tuning are crucial. LLM outputs are often calibrated using reinforcement learning from human feedback (RLHF) to avoid unsafe or irrelevant answers (e.g. framing medical advice conservatively) [20].

### C. Responsible AI in Healthcare

Recent WHO guidance [10] has warned against premature deployment of LLMs in healthcare, highlighting risks of hallucinated, biased, or incomplete outputs that can harm patients, especially in consumer-facing or under-resourced settings. The 2024 WHO ethics framework for large multimodal models (LMMs) expands these concerns, calling for rigorous evaluation, informed consent, representative datasets, transparency, and safeguards against algorithmic harm.

These principles are further contextualized in the systematic review by [9], which synthesizes over 100 studies to identify practical initiatives aligned with responsible AI adoption. These include ensuring fair data governance, explainability in decision-making, inclusive stakeholder engagement, and ongoing monitoring. Despite these calls for ethical oversight, most LLM-based health systems lack embedded mechanisms for real-time accountability or explainability. Few existing digital twin implementations address social bias or data privacy while synthetically generating EHR [4]. However, they have not provided a holistic approach to map standard ethical guidelines (e.g., WHO guidelines) for the use of LLMs in healthcare. To fill this gap, we design RHealthTwin to directly operationalize WHO's six principles within a responsible prompt engine that regulates behavior, ensures fairness, and supports longitudinal digital twin alignment with evolving user needs. We highlight the principle, description from WHO, associated risks (from LLM deployments), corresponding safeguards, and how each is operationalized in the ResponsibleHealthTwin framework in Table II.

## III. METHODOLOGY

The proposed framework leverages a digital twin architecture to improve consumer health applications through personalized, context-aware, and multimodal interactions. As illustrated in Figure 1, RHealthTwin system integrates user data with advanced AI-driven modules to create a digital representation (digital twin) of well-being assistant, allowing customized health interventions, risk identification and scenario generation. The framework is designed to ensure autonomy, inclusivity, transparency aligning with human well-being principles. To fomalize, The RHealthTwin framework processes multimodal inputs $M$ via the Responsible Prompt Engine to generate structured prompts $P = \mathsf{F}_{\text{RPE}}(M, \theta_{\text{RPE}})$. These prompts guide an LLM in healthcare care $\mathsf{H}_{\text{LLM}}(P) \rightarrow D_{\text{twin}}$ to produce results, with parameters updated through $\theta_{t+1} = \theta_t + \theta \nabla_\theta \mathsf{L}(\Delta_{feedback}, \mathsf{E})$, where $\mathsf{E}$ enforces ethical

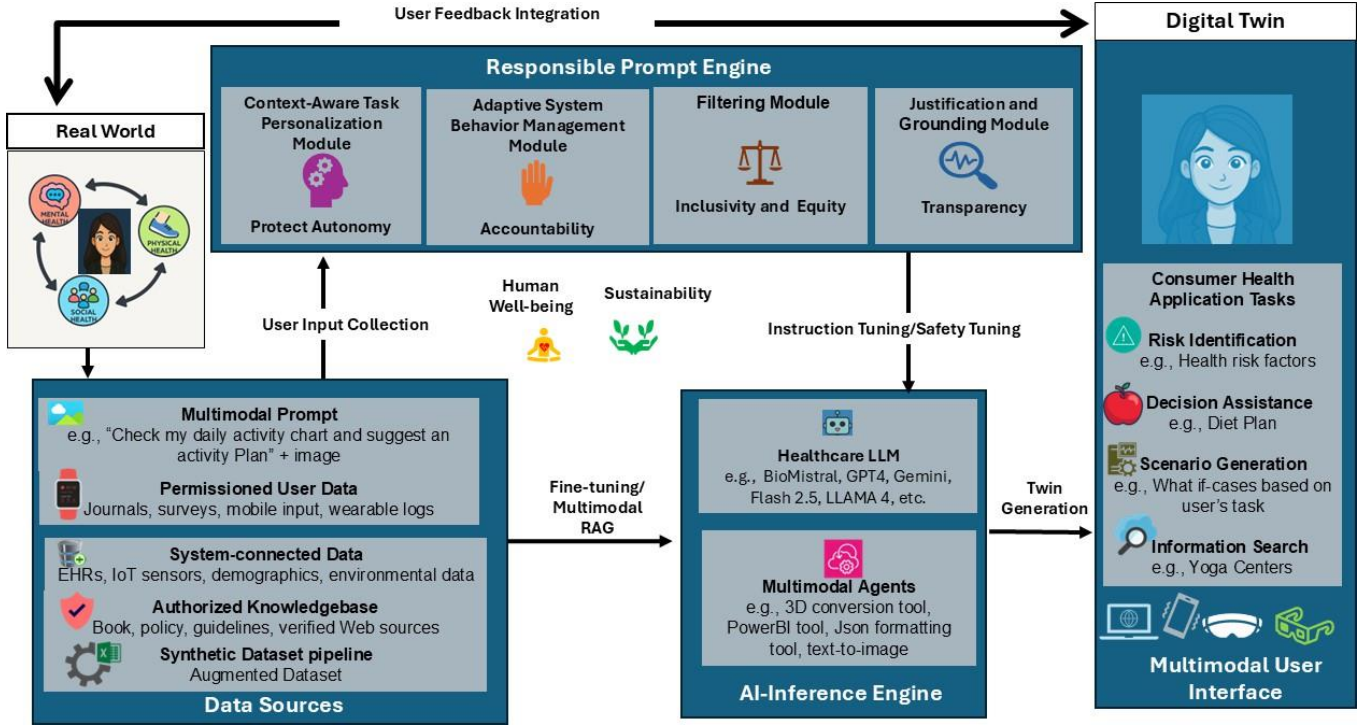

Fig. 1. Proposed RHealthTwin Framework

alignment and $\Delta_{feedback}$ refines results through user interactions.

### A. Example Scenario

To illustrate how our framework builds a personalized well-being digital twin, let us consider the case of a 31-year-old working mother who queries: *"Check my weekly activity log and suggest how I can reduce fatigue without compromising productivity."*. As illustrated in Fig. 2, the RHealthTwin system first collects multimodal inputs, such as a fitness app screenshot showing low step count, high screen time, and elevated evening heart rate, along with self-reported fatigue and irregular meals. The Context Module synthesizes these signals to define her health goal—fatigue reduction while preserving productivity. The Instruction Module configures the LLM to act as a safe, ethical well-being assistant, avoiding unverified or high-risk recommendations. These modules are composed by the RPE, which integrates few-shot reasoning examples, safety filters, and credible justifications such as CDC sleep guidelines (See Fig. 3). The LLM processes this structured prompt to generate four personalized outputs: (1) decision guidance (e.g., yoga routine and fixed sleep schedule), (2) risk alerts (e.g., elevated room temperature disrupting sleep), (3) scenario simulation (e.g., expected REM improvement), and (4) contextual search results (e.g., nearby yoga classes). The user interacts with these suggestions through a multimodal interface, endorsing helpful advice and skipping less relevant ones (See Fig. 8). This feedback is looped into her DT well-being assistant, enabling continual refinement and alignment with her evolving well-being needs. This end-to-end workflow demonstrates how our framework responsibly generates a dynamic and reliable well-being assistant customized for real-world health support.

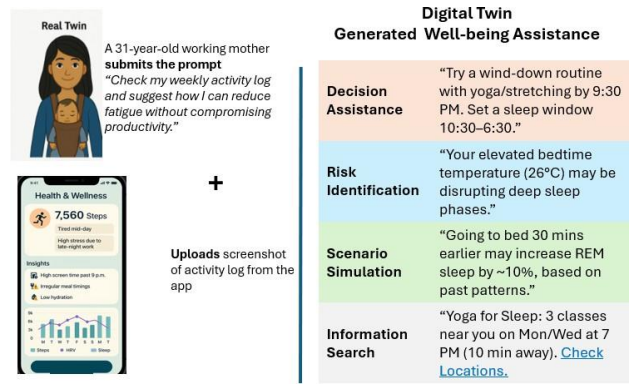

Fig. 2. Example Use-Case Scenario for Proposed RHealthTwin Framework

### B. Data Sources and Integration

The RHealthTwin framework integrates two primary data streams to generate personalized digital twin outputs: (1) *user-provided data* (direct inputs) and (2) *connected resources* (external data authorized by the user). User directly inputs natural-language queries or commands (optionally accompanied by images or recorded data), and the system also ingests permissioned personal health data (e.g., logs from wearable devices, EHRs, self-report surveys, journals). Envi-ronmental and contextual information (e.g., location, weather, daily schedule, sleep and activity logs) is also included to situate recommendations. For example, a user query "Help me improve my sleep routine" might come with a sleep-tracker chart image, last night's heart-rate data, and a note that the user has been working late. By integrating diverse data modalities – text, images, time-series, etc. RHealthTwin mimics real-world clinical scenarios where information is inherently multimodal. In addition, RHealthTwin has access to an authorized knowledge base (clinical guidelines, textbooks, vetted web sources) and synthetic data pipelines to augment learning. All data use is strictly permissioned by the user, protecting privacy while enabling the twin to form an up-to-date, contextual health profile. We formalize the input space, $\mathsf{M}$ as:

$$\mathsf{M} = \mathsf{M}_{user} \cup \mathsf{M}_{connected}, \qquad (1)$$

where:

- $\mathsf{M}_{user} = \{x_{text}, x_{image}, x_{self\text{-}report}, \ldots\}$ includes explicit user inputs (e.g., free-text queries, uploaded images, or manual logs).
- $\mathsf{M}_{connected} = \{x_{EHR}, x_{wearable}, x_{knowledge}, \ldots\}$ aggregates permissioned external data (e.g., EHR records, wearable sensor streams, and medical knowledge bases).

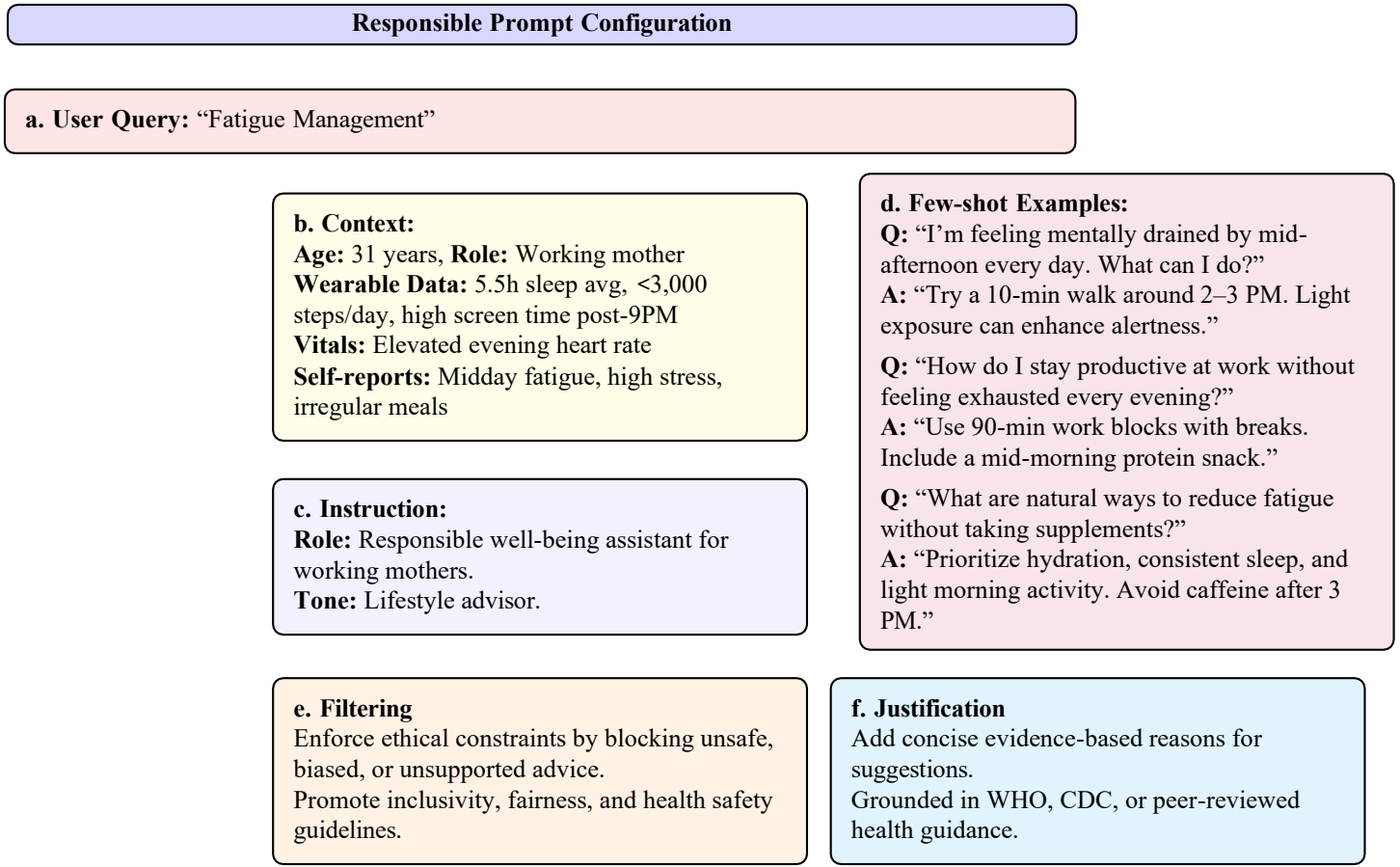

Fig. 3. An Example of LLM-based Fatigue Reduction with Ethical Constraints and User Context.

The RPE further fusion these inputs via:

$$\tilde{\mathsf{M}} = \mathsf{F}_{preprocess}(\mathsf{M}, \theta_{constraints}), \qquad (2)$$

where $\theta_{constraints}$ enforces the input to be aligned with ethics standards. RPE retrieves the context from user query and dynamically generates user prompt and system instructions with necessary instructions, fewshot examples, and filters (See Fig. 3). We detail the proposed approaches for RPE as follows.

### C. Responsible Prompt Engine (RPE)

At the core of the RHealthTwin framework lies the RPE, which governs the ethical and effective use of LLM-based inference in sensitive consumer health applications. The RPE transforms an unconstrained user input—typically a short, natural-language query into two well-defined prompt structures that align with responsible AI principles:

- **System Instruction:** Defines the assistant's behavior, including its role, tone, ethical filters, and format style via few-shot examples.
- **User Prompt:** Embeds the user's health-related goal, contextual profile, and justification constraints to guide personalized and explainable generation.

Unlike conventional LLM interfaces that rely on user-controlled system prompts, our proposed RPE enforces a structured inference mechanism through dynamic slot tagging and semantic template parsing. We describe this approach, the description of each modules in RPE and algorithms to perform this processing. In this study our target slots are defined as in Table III.

TABLE III
SLOT DEFINITIONS USED IN RESPONSIBLE PROMPT ENGINE

| Slot | Description |
|------|-------------|
| UQ | Extract user's well-being query in one sentence. |
| CP | Identify user health context such as age, sleep patterns, lifestyle factors, or recent observations. |
| J | Responsible health principles that should guide the generated output (e.g., evidence-based, non-prescriptive). |
| ROLE | Role of the assistant such as "health advisor" or "wellness coach" to simulate appropriate expertise. |
| TONE | Preferred response tone such as friendly, neutral, or encouraging. |
| FILT | Content filtering rules and safety constraints to avoid harmful, biased, or inappropriate suggestions. |
| FE | Few-shot example pairs (query and response) for in-context learning and demonstration. |

This innovative approach ensures consistent and contextually relevant outputs by systematically extracting and adapting key components of user input. We outline the architecture of RPE, detailing each module and the algorithms that facilitate this processing as follows.

*1) Context-Aware Prompt Module:* This component augments the user's query with explicit goals and background context. It identifies the user's intent (e.g. "improve sleep"), health history, current state, and preferences, and injects this into the prompt. By grounding the task in the user's personal context and objectives, it protects user autonomy and ensures the output aligns with what the individual truly wants. For instance, it might append details like known insomnia issues or caffeine intake to the query.

We define the context-aware prompt $P_{\text{user}}$ as the composition of the user query $\text{UQ} \in \mathsf{Q}$ and contextual parameters $\text{CP} \in \mathsf{C}$, both extracted from $M$ using structured slot-based transformations:

$$\text{UQ} = \mathsf{F}_{\text{slot}}(M; \theta_{\text{UQ}}) \quad (3)$$

$$\text{CP} = \mathsf{F}_{\text{slot}}(M; \theta_{\text{CP}}) \quad (4)$$

$$P_{\text{user}} = \mathsf{G}_{\text{compose}}(\text{UQ}, \text{CP}) \quad (5)$$

The operator ($\mathsf{F}_{\text{slot}}$) maps the input ($M$) to the slot values using template-guided semantic extraction governed by slot-specific parameters ($\theta$). The function ($\mathsf{G}_{\text{compose}}$) injects the extracted context into the user query, forming a personalized prompt:

$$P_{\text{user}} = \begin{cases} \mathsf{G}_{\text{compose}}(\text{UQ}, \text{CP}), & \text{if } \text{CP} \neq \emptyset \\ \text{UQ}, & \text{otherwise} \end{cases} \quad (6)$$

This ensures consistent structure in prompt composition and prevents silent failures when context is missing. As illustrated in Figure 2, user inputs a prompt and screenshot from fitness app, which is denoted as $M$ in the equation. The UQ and CP for this example is depicted in Fig. 3 (a).

*2) System-Adaptive Behaviour Module:* We define the following slot values extracted from $M$ using structured slot-based templates for the system behaviour parameters defined as:

$$\begin{aligned} \mathsf{P}_{\text{systemBehavior}} &= \mathsf{F}_{\text{slot}}(M; \theta_{\text{ROLE}}) \cup \mathsf{F}_{\text{slot}}(M; \theta_{\text{TONE}}) \\ &\cup \mathsf{F}_{\text{slot}} \cup \mathsf{F}_{\text{slot}}(M; \theta_{\text{FE}}) \end{aligned} \quad (7)$$

Where ($\theta_{\text{ROLE}}$, $\theta_{\text{TONE}}$, $\theta_{\text{FE}}$, denotes prompt templates for extracting the to set the role (e.g., well being assistant), tone (e.g., engaging), fewshot examples (e.g., output examples for input examples in JSON format), respectively. Figure 3(c) and Figure 3(d) shows examples for these slots.

*3) Filtering Module:* This module enforces fairness, privacy, and safety by injecting explicit constraints and content filters into the prompt. The constraints target addressing accountability and appropriateness in critical healthcare contexts. Filtering module embeds such controls (e.g., always advising to consult a doctor for high-risk issues) to align the system behavior with professional standards.

We define the following slot values extracted from $M$ using structured slot-based templates for the filtering module is defined as:

$$\mathsf{P}_{\text{filter}} = \mathsf{F}_{\text{slot}}(M; \theta_{\text{FILT}}) \quad (8)$$

Where $\theta_{\text{FILT}}$ denotes filters (e.g., no medication suggestion) necessary to possible harmful content that may present in the hallucinated output as examplified in Figure 3(e) This slot defines content constraints (e.g., safety, legality, bias prevention) injected into the system instruction. The filtering output is composed with other instruction modules.

*4) Justification Module:* This module retrieves relevant supporting data to enable explainability. It queries authorized clinical knowledge bases or the user's own health records for evidence (e.g., established sleep hygiene guidelines when answering a sleep query). By appending pertinent facts or citations to the prompt, it grounds the LLM's reasoning in verifiable information. This aligns with evidence-based approaches to frame the response as transparent and justifiable for trustworthy well-being interventions.

The justification constraint is extracted using its assigned prompt template:

$$\mathsf{P}_{\text{justification}} = \mathsf{F}_{\text{slot}}(M; \theta_{\text{JUST}}) \quad (9)$$

This slot enforces evidence-driven explanation requirements (e.g., cite reasoning steps, support with guidelines). It is appended to the user prompt $\mathsf{P}_{\text{user}}$.

*5) Slot Tagging and Template-Based Extraction:* In safety-critical domains such as healthcare, unconstrained user instructions can inadvertently bypass ethical filters, leading to biased or unsafe outputs [9] [10]. To address this, we formalize slot-value extraction as a controlled natural language generation task with explicit structural boundaries, enhancing reproducibility, interpretability, and safety as following.

Each slot $s \in \mathsf{S}$ is associated with a predefined template $\theta_s$, which wraps the user input with a natural language instruction and structural tags (e.g., <UQ>, </UQ>). The model output is then post-processed using span-based information extraction, a common technique in sequence labeling and constrained text generation. The value of each slot is computed as:

$$\text{SlotValue}_s = \text{Span}_{\text{StartTag},\text{EndTag}} (\mathsf{F}_{\text{LLM}}(\theta_s(M))) \quad (10)$$

where $\mathsf{F}_{\text{LLM}}$ denotes the generative model inference function and $\theta_s(M)$ is the templated prompt generated by injecting the user input $M \in \mathsf{M}$ into the slot-specific instruction.

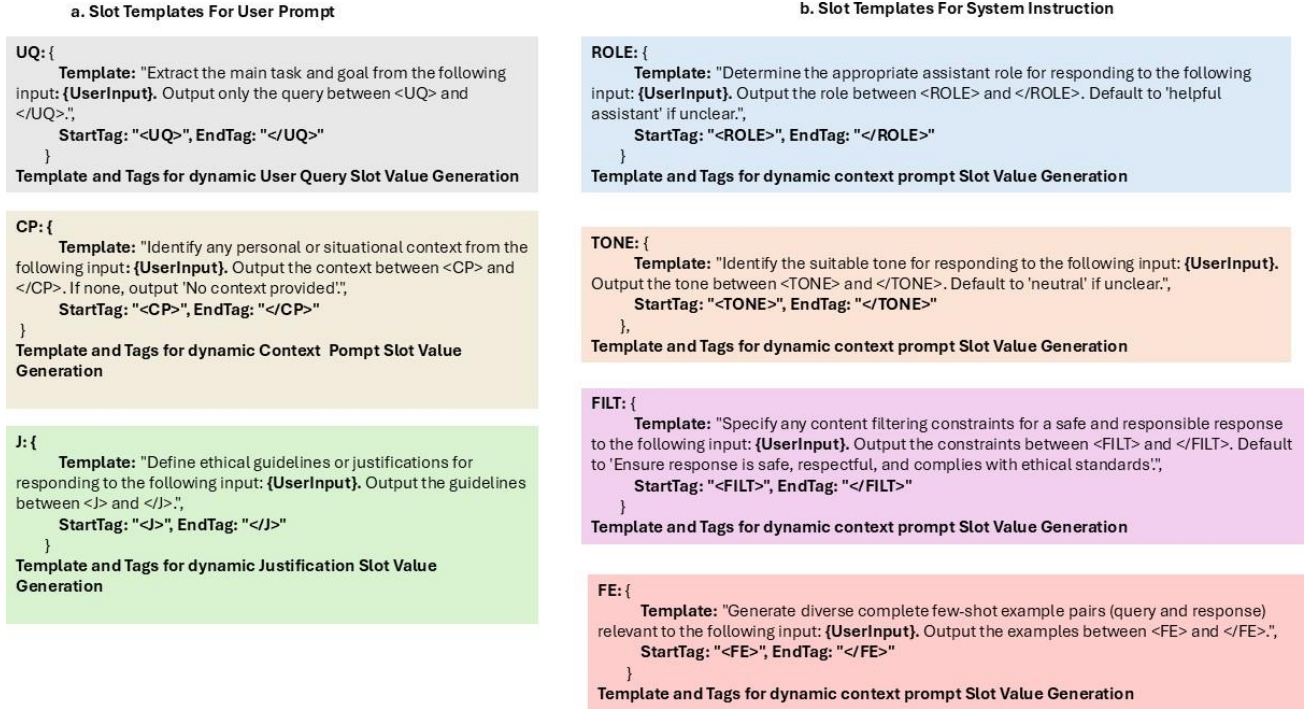

Fig. 4. Structured Slot Template Schema for Responsible Prompt Generation

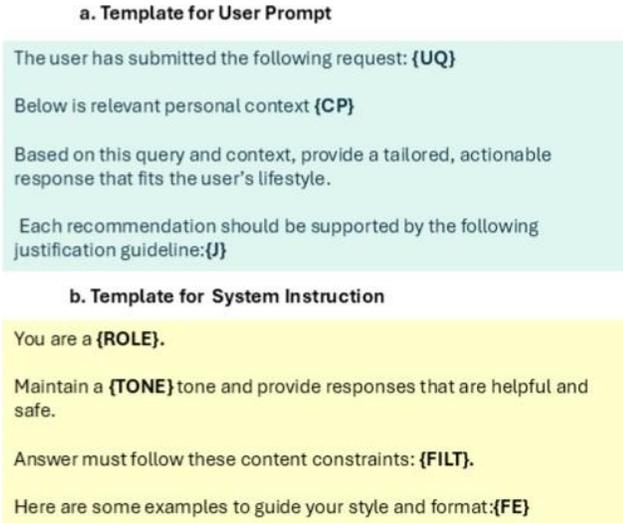

Fig. 5. Structured prompt template for responsible well-being response generation. **(a)** User Prompt Template combines the user query (UQ), personal context (CP), and justification guideline (J) to guide tailored response generation. **(b)** System Instruction Template configures the assistant's behavior, specifying role (ROLE), tone (TONE), safety constraints (FILT), and illustrative few-shot examples (FE) to ensure ethical and compliant output.

The extracted slot values are then composed into the final `UserPrompt` and `SystemInstruction`, which are forwarded to the domain-specialized health LLM for responsible response generation.

Algorithm 1, facilitates the process of our proposed adaptive slot-based responsible prompt engineering to dynamically construct the final user prompt and system instruction, which is key to inference the healthcare LLM. The first part of the algorithm focus on retrieving the slots using the predefined templates as demonstrated in Figure 4. The second part of the algorithm focus on structuring the user prompt and system instruction into a predefined slot based template as illustrated in Figure 5.

### D. AI Inference Engine

The AI Inference Engine in the RHealthTwin framework comprises two core components: (1) a Healthcare LLM and (2) a Multimodal Agent.

The **Healthcare LLM** can be any instruction-tuned large language model suitable for health and well-being applications, such as GPT-4, Gemini Flash 2.5, LLaMA 4, or Qwen-V2. It may also refer to a domain-adapted variant of these models fine-tuned on specific healthcare tasks. For instance, in one of our ongoing studies, LLaMA has been fine-tuned on

**Algorithm 1** Adaptive Slot Based RPE
  **Input:** UserInput
  **if** UserInput is image OR audio **then**
    TextInput ← MultimodalLLM.convertToText(UserInput)
  **else**
    TextInput ← UserInput
  **end if**
  Initialize SlotTemplates with predefined templates and tags

  Slots ← {}
  **for** each SlotName in SlotTemplates **do**
    PromptText ← Replace(SlotTemplates[SlotName].Template, {TextInput}, TextInput)
    SlotOutput ← MultimodalLLM.generate(PromptText)
    StartTag ← SlotTemplates[SlotName].StartTag
    EndTag ← SlotTemplates[SlotName].EndTag
    StartIndex ← FindIndex(SlotOutput, StartTag) + Length(StartTag)
    EndIndex ← FindIndex(SlotOutput, EndTag)
    **if** StartIndex ≥ Length(StartTag) AND EndIndex > StartIndex **then**
      Slots[SlotName] ← Substring(SlotOutput, StartIndex, EndIndex)
    **else**
      Slots[SlotName] ← ""
    **end if**
  **end for**
  UserPrompt ← CREATEUSERPROMPT(Slots)
  SystemInstruction ← CREATESYSTEMINSTRUCTION(Slots)
  **Return:** (UserPrompt, SystemInstruction)

**Algorithm 2** Responsible AI Inference
  **Input:** UserInput, MultimodalLLM, SessionState, UseRAG, UseWeb, UseAgent
  **Output:** Final LLM Response
  (UserPrompt, SystemInstruction) ← EXTRACTPROMPTSLOTSCLEANLY(UserInput, MultimodalLLM)
  GroundingSnippets ← [ ], AgentResults ← [ ]
  **if** UseRAG **then**
    Keywords ← EXTRACTKEYWORDS(UserPrompt)
    GroundingSnippets ← MULTIMODALRAGSEARCH(Keywords, MaxResults=5)
  **else if** UseWeb **then**
    Keywords ← EXTRACTKEYWORDS(UserPrompt)
    GroundingSnippets ← PERFORMWEBSEARCH(Keywords, MaxResults=5)
  **end if**
  **if** UseAgent **then**
    AgentTasks ← IDENTIFYAGENTTASKS(UserPrompt, SystemInstruction)
    **for** each Task in AgentTasks **do**
      **if** Task.Type = "SendEmail" **then**
        EmailContent ← GENERATEEMAILCONTENT(UserPrompt, GroundingSnippets, MultimodalLLM)
        AgentResults.append(SENDEMAIL(EmailContent))
      **else**
        AgentResults.append(EXECUTEAGENTACTION(Task))
      **end if**
    **end for**
  **end if**
  Messages ← [ ]
  Messages.append({Role: "system", Content: SystemInstruction})
  **if** GroundingSnippets is not empty **then**
    Evidence ← "[Retrieved Evidence] " + JOIN(GroundingSnippets, "\n")
    Messages.append({Role: "assistant", Content: Evidence})
  **end if**
  Messages.append({Role: "user", Content: UserPrompt})
  Response ← MultimodalLLM.GenerateChat(Messages)
  **if** AgentResults is not empty **then**
    Response ← Response + "\n[Agent Actions]\n" + JOIN(AgentResults, "\n")
  **end if**
  SessionState["chat_history"].append({UserPrompt, SystemInstruction, LLMResponse: Response})
  **Return:** Response

a curated mental health dialogue dataset to generate context-aware and empathetic responses.

The **Multimodal Agent** extends the capabilities of the Healthcare LLM by enabling response augmentation and real-world automation. It enhances accessibility by transforming LLM-generated outputs into actionable formats, such as converting structured JSON responses into user-friendly messages, reminders, or visual summaries. The agent can also interface with third-party applications or APIs to trigger follow-up actions based on AI responses.

In the RHealthTwin framework, **Data Sources** supply input for both fine-tuning and RAG operations. RHealthTwin supports a dedicated fine-tuning pipeline as well as an optional multimodal RAG setup to enhance the performance and contextual grounding of the Healthcare LLM. This enables the Healthcare LLM to retrieve relevant supporting information from internal knowledge bases or external sources, grounding its outputs with verifiable context. For example, in [21] such as LLMTherapist, a multimodal RAG-based architecture has been employed to support mental health counseling by integrating user history, affective cues, and clinical literature. Similarly, RHealthTwin allows grounding snippets to be injected into the system prompt to enhance factuality and therapeutic alignment.

**Algorithm 3** Update Slots From Feedback
1: **Input:** FeedbackInput, SessionState, MultimodalLLM
2: **Output:** Updated SlotTemplates
3: **if** SessionState.FeedbackFlag = 1 **then**
4:     FeedbackIntent ← MultimodalLLM.Generate("Extract semantic intent from: " + FeedbackInput + "\nOutput: <INTENT>...</INTENT>")
5:     Intent ← ExtractSubstring(FeedbackIntent, \<INTENT>", \</INTENT>")
6:     Last ← SessionState["chat_history"][-1]
7:     SlotTemplates ← UpdateSlotTemplates(Intent, Last["slot_templates"])
8:     Last["slot_templates"] ← SlotTemplates
9:     SessionState["chat_history"][-1] ← Last
10:    **return** SlotTemplates
11: **end if**

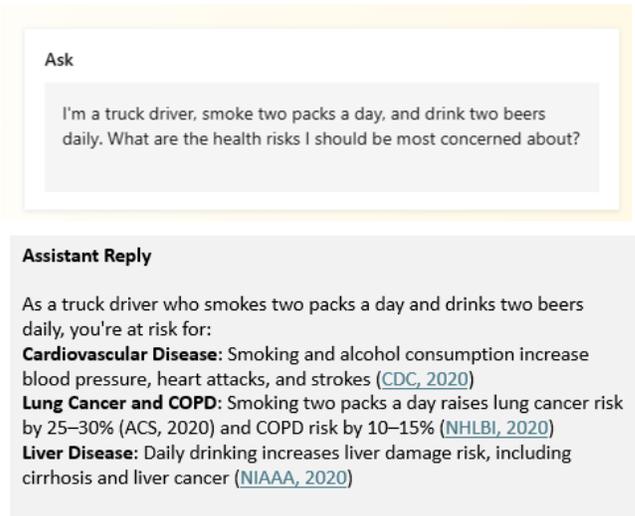

Fig. 6. Digital Twin assessment of personalized health risks for a truck driver using lifestyle indicators (smoking, alcohol use).

As depicted in Figure 1, the **RPE** modules enable instruction tuning and safety tuning to the AI-inference engine. The AI-Inference Engine consumes the `UserPrompt` and `SystemInstruction` produced by the Slot-Based RPE (see Algorithm 1). These inputs are passed into the session-level inference algorithm (Algorithm 2). The refined prompts from the RPE feed into the AI-Inference Engine. In practice, we employ instruction tuning and safety tuning to align the LLM with clinical needs. Because careful prompt and instruction engineering allows open source LLMs to rival fine-tuned models on medical benchmarks [22]. We describe these processes as follows.

*1) Instruction Tuning:* In the *System Instruction construction step* of Algorithm 2, instruction tuning is applied at inference time by injecting role, tone, and task directives into the assistant's context. This method enables behavior alignment without retraining the base model. As supported by recent literature [22], instruction tuning yields more user-aligned and explainable outputs while avoiding the high compute and dataset requirements of full fine-tuning.

*2) Safety Tuning:* Safety constraints, derived from the `FILT` and `J` slots, are injected into the system instruction during the same step. This design ensures that the LLM adheres to ethical guardrails and avoids hallucinations or medically inappropriate suggestions. The inclusion of justification directives also improves the transparency and trustworthiness of the model's recommendations, particularly in sensitive healthcare applications.

*3) Multimodal RAG:* The optional *Evidence Retrieval step* in Algorithm 2, enhances factual grounding through keyword-based search using either a local RAG pipeline or external web sources. When enabled, relevant snippets are prepended to the assistant context prior to generation. This allows the LLM to reference external evidence without requiring internal memorization of medical facts. Injecting retrieved information into the assistant message improves the factuality and traceability of the response, especially for health-related queries.

### E. DT Well-being Assistant

The DT in the RHealthTwin framework (see Figure 2) functions as a dynamic, personalized health companion that generates individualized well-being insights based on multi-modal user input and responsible prompt engineering. Once generated via the AI inference engine, the the DT operates across multiple functional dimensions, aligning with established digital twin classifications as follows.

- **Information Twin:** The DT continuously aggregates and analyzes data from various sources, including wearable devices, self-reported inputs, and environmental sensors. This integration offers a comprehensive view of the user's current health status, enabling the identification of latent risk factors such as stress-induced hypertension or irregular sleep patterns. Figure 6, illustrates an example where smoking and alcohol-related risks are summarized for a truck driver based on known consumption levels.
- **Predictive Twin:** Leveraging historical data and advanced analytics, the DT simulates potential future health scenarios. For instance, it can model the impact of behavioral changes, such as adjusting sleep schedules or dietary habits, on health outcomes like fatigue levels or cardiovascular risk. This predictive modeling empowers users to anticipate the consequences of lifestyle modifications. Figure 7, shows how occupational routines (e.g., truck driving) influence activity distribution and inform future planning.
- **Action Twin:** Beyond monitoring and prediction, the DT provides personalized, actionable recommendations customized to the user's goals and context. These interventions may include suggestions for physical activity, nutritional adjustments, or stress management techniques. By aligning recommendations with real-world constraints and user preferences (as shown in Figure 7), the DT supports search and documentation of resources (e.g.

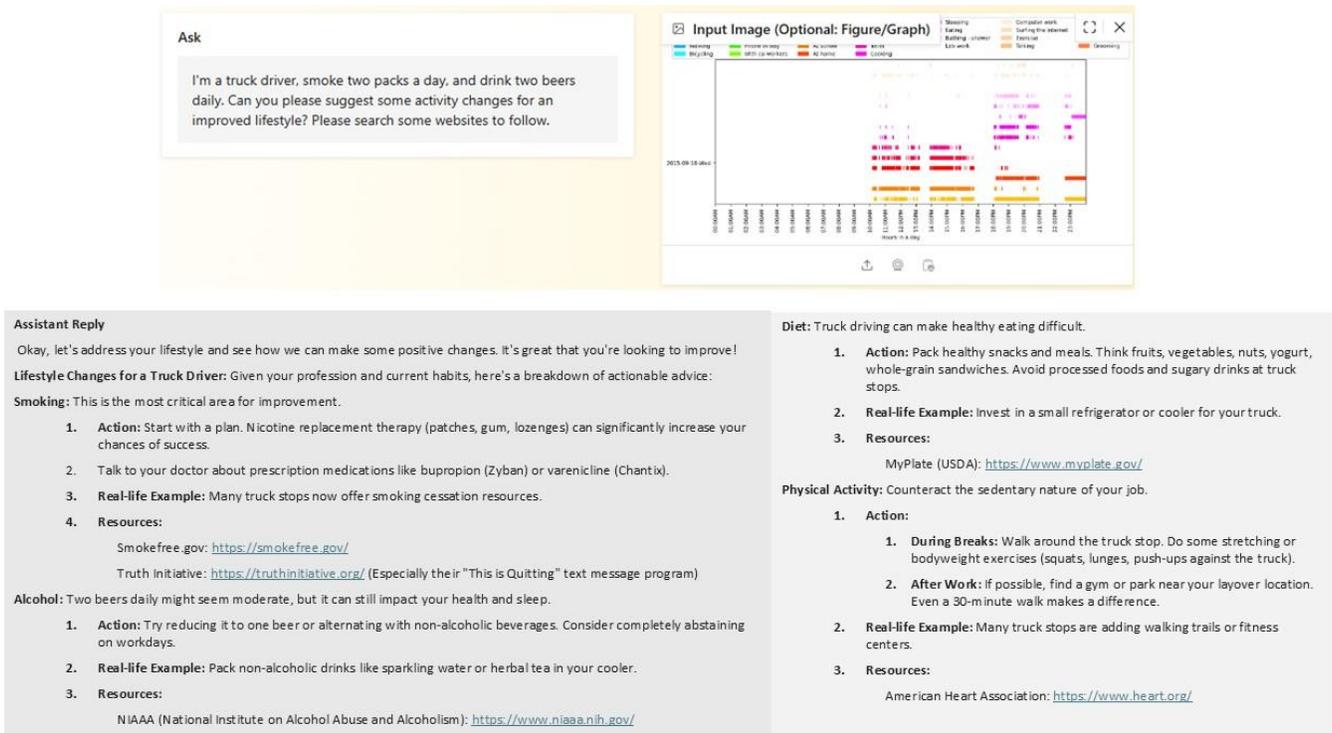

Fig. 7. Context-aware Digital Twin recommending lifestyle interventions based on profession-specific constraints and time-activity patterns.

useful websites).

The RHealthTwin exemplifies an intelligent health management system. By identifying latent risks (e.g., prediabetes, insomnia patterns), facilitating informed decision-making, and supporting continuous learning, it adheres to the latest standards in health digital twin frameworks. Its iterative refinement through user feedback ensures sustained accuracy and relevance in promoting user well-being. We detail the close-loop feedback mechanism in the following section.

To refine responses and continuously personalize interactions, we integrate a feedback mechanism that processes user reactions—submitted as text or structured inputs (like/dislike), via a designated feedback field in the user interface. When activated, this feedback loop analyzes the semantic intent behind user input and updates the responsible prompt generation pipeline. The process ensures that subsequent prompt construction (e.g., slot templates for query interpretation, role, tone, or filtering) reflects the user's evolving preferences and concerns.

### F. Feedback Integration and Adaptive Message Tuning

In RHealthTwin framework feedback is captured via a specific feedback field in the multimodal user interface. This enables users to comment on or rate previous AI outputs. Both structured inputs (like/dislike buttons) and unstructured text (e.g., "This advice feels too generic") are supported. While this feedback loop facilitates user-centric personalization and system accountability, its true impact relies on stakeholder collaboration. As shown in Figure 8, the user interacts with the system through a feedback field in the interface. They express their experience using free-text input or structured responses (e.g., like/dislike). This feedback is semantically parsed to extract positive or negative sentiment, key behavior cues (e.g., "yoga," "early dinner"), and intent signals (e.g., "keep reminding me"). The system integrates this contextual information into the real user Profile, modifies the system instruction accordingly, and updates future prompts to reflect the user's preferred routines. In this way, the system adapts to evolving user preferences and reinforce positive habits through adjusted system behavior in future sessions.

The Algorithm 3 is triggered only when the "FeedbackFlag" is set. It semantically interprets the user's feedback, such as "avoid strict diets" or "prefer relaxing evening plans" and updates the relevant slot templates (e.g., Constraint Preferences, Filtering rules). These updated templates are stored in the session state and automatically influence the next AI response by modifying how user input is parsed and interpreted during the next Algorithm 1 call.

Algorithm 3, enables user feedback adaptation by updating the internal slot template structure of the Responsible Prompt Engine based on semantic user feedback. When the FeedbackFlag is active, the system treats FeedbackInput as a prompt and passes it to the MultimodalLLM, which extracts the user's implicit intent using tag-based delimiters (e.g., <INTENT>... </INTENT>). This extracted intent may include preferences (e.g., "prefer yoga"), aversions (e.g., "avoid caffeine-related advice"), or stylistic instructions (e.g., "simpler tone"). The updated slot templates are stored within the latest entry of the

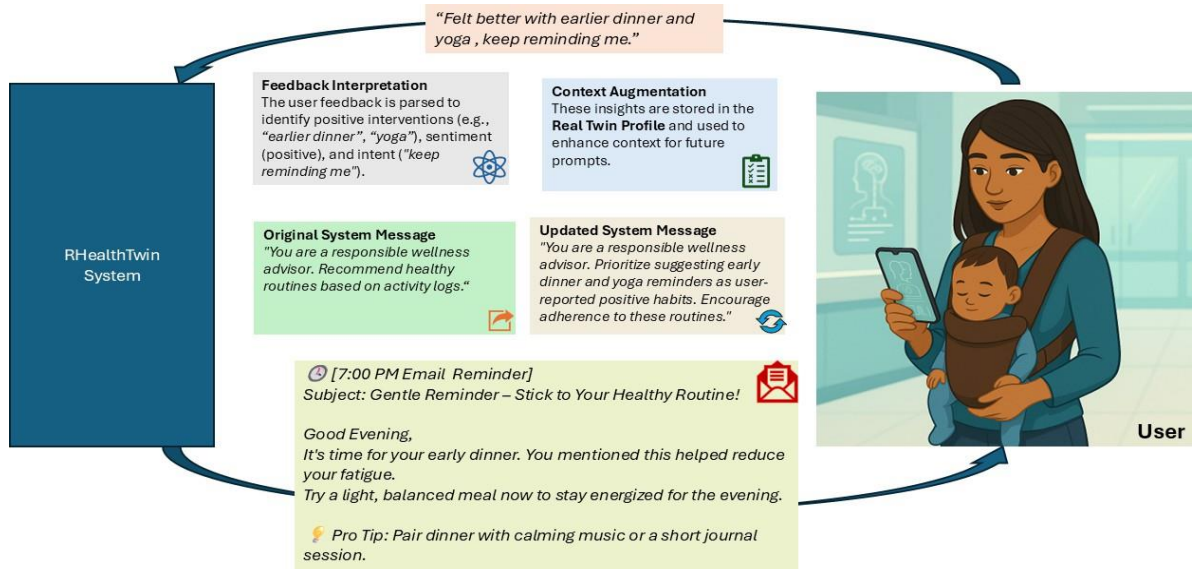

Fig. 8. Personalized wellness feedback loop to address user feedback to refine responses.

session's chat history, ensuring continuity across interaction sessions and allowing longitudinal learning over time.

While this feedback loop improves personalization, it alone does not guarantee clinically optimal outcomes. To realize its full potential, collaboration with stakeholders. Such as healthcare professionals, caregivers, and ethical reviewers is essential. They ensure the interpretations and updates remain safe, equitable, and aligned with domain standards, especially in high-stakes health applications.

## IV. Experiment

In this section, we present the dataset, experiment and result analysis as followings. While the complete system is modular and extensible, this study focuses primarily on evaluating the Responsible Prompt Engine. Specifically, its ability to generate structured, ethical, and personalized prompts that improve response quality from large LLMs. Given the expansive nature of the framework, we restrict the current evaluation to:

- The impact of slot-based prompting on LLM output quality
- Comparison with standard prompting paradigms (zero-shot, few-shot, instruction-tuned) to understand effect of RPE for safe, transparent, and grounded generated output.

### A. Dataset

We conduct experiments across four benchmark datasets spanning diverse health domains, including mental health, clinical QA, nutrition planning, and lifestyle monitoring. In Table IV, we summarize the dataset properties, use case categories, and sample sizes.

**MentalChat16k (16,113 dialogues)** consists of synthetic and anonymized therapist–patient conversations covering topics such as depression, anxiety, and emotional distress. We use it to assess RHealthTwin's ability to generate empathetic and safe responses in mental health support scenarios. **MTS-Dialog v3 (1,701 dialogues)** contains structured doctor–patient consultations paired with clinical summaries. It serves as a testbed for evaluating RHealthTwin's capability to generate clinically coherent and role-appropriate responses under diagnostic or procedural queries. **NutriBench v2 (11,857 entries)** provides human-verified meal descriptions annotated with macronutrient content. This dataset is used to assess how well RHealthTwin delivers accurate and personalized dietary recommendations grounded in nutritional standards. **SensorQA (5,600 QA pairs)** includes questions derived from long-term wearable sensor data. It supports evaluating RHealthTwin's ability to interpret multimodal behavioral logs and offer personalized lifestyle advice, such as sleep, activity, or hydration guidance.

### B. Test Prompts

To enable responsible, personalized dialogue across datasets that lacked structured prompts, we synthetically generated two sets of prompts per instance: one simulating the patient side and another from the perspective of a care provider (e.g., doctor, nutritionist, or AI health coach). For each sample, we used the structured attributes (e.g., symptoms, nutrition labels, sensor data) to generate a realistic, context-rich patient query, alongside a corresponding provider question to initiate or follow-up on a relevant health conversation. This setup reflects real-world human-AI interactions in both reactive and proactive use cases. For example, in the NutriBench v2 dataset: *Patient Prompt: "I'm a vegetarian trying to manage my iron levels. What can I cook with lentils and spinach that's also low in carbs?"* Care Provider Prompt: *"Given this meal description, are there any missing nutrients or contraindicated combinations for a patient with low iron and mild anemia?"*.

TABLE IV
BENCHMARK DATASETS WITH WELL-BEING USE CASES IN RESPONSIBLEHEALTHTWIN EVALUATION

| Dataset | Domain | Simulated Role | Modality | Total Size | Subset | Well-being Use Case |
|---|---|---|---|---|---|---|
| **MentalChat16k** | Mental health Q&A | Therapist ↔ Patient | Text (dialogue) | 16,113 | 2,400 | Mental health journaling and emotional support |
| **MTS-Dialog v3** | Clinical task simulation | Doctor ↔ Patient | Text (clinical query) | 3,603 | 540 | Symptom triage, diagnostics, clinical instruction |
| **NutriBench v2** | Nutrition & safety | Nutritionist ↔ Patient | Tabular → Text | 9,200 | 1,380 | Dietary recommendation and nutrition planning |
| **SensorQA** | Lifestyle coaching | AI Coach ↔ Wearables | Screenshot + Text Prompt | 1,200 | 180 | Sleep, activity, and habit balancing suggestions |

TABLE V
SYNTHETIC EXAMPLES OF PATIENT AND CARE PROVIDER PROMPTS ACROSS DATASETS

| Dataset | Patient-Side Prompt | Care Provider-Side Prompt |
|---|---|---|
| MentalChat16k | I've been feeling very anxious lately and can't seem to calm down, even when I'm trying to relax. | How can we best support a patient who is presenting with persistent anxiety and emotional distress? |
| MTS-Dialog v3 | I've had a persistent cough for three days, along with mild fever and fatigue. Should I be concerned? | Does the patient's cough pattern and associated symptoms indicate a potential respiratory infection requiring further evaluation? |
| NutriBench v2 | I'm a vegetarian trying to manage my iron levels. Could you suggest some plant-based foods that can help? | Given this meal description from a vegetarian patient, are there any missing nutrients or adjustments needed to improve iron intake? |
| SensorQA | I've uploaded a screenshot of my fitness tracker summary. I only slept four hours and didn't meet my step goal yesterday. Should I make any changes? | Based on the uploaded wearable summary showing reduced sleep and low activity, what adjustments should be recommended to improve overall well-being? |

In Table V, we represent the example test prompts used in our experiment. The prompts used in these datasets are available in RHealthTwin's open-source repository. [5].

### C. Selected Models

We selected a diverse range of large language models to systematically assess how RPE handles structured prompt generation, and responsibility enforcement across both general-purpose and healthcare-specialized LLMs. For multimodal understanding and grounded inference, we included **LLAMA 4**[6], **Gemini 2.5 (Flash and Pro)**[7], and **GPT-4**[8], aligning with use cases from studies like HealthLLM [22] and DT-GPT [6]. We also incorporated domain-specific biomedical models such as **BioMistral-7B**[9] and **Asclepius-7B**[10]. To benchmark against widely used open models in prior work such as Twin-GPT [4] and PsyDT [23], we additionally tested **LLaMA3-8B-Instruct**[11], **Qwen2-7B-Instruct**[12], **Qwen VL**[13], **Mistral-7B**[14], and **GPT-3.5**[15].

### D. Responsible Prompt Generation

To ensure a balanced trade-off between coverage and computational feasibility, we sample approximately 15% of each benchmark dataset for evaluation. This sampling rate is inspired by prior findings in the HealthLLM study [22], which demonstrated that fine-tuning only 15% of health-specific datasets can already yield significant performance improvements over zero-shot baselines across diverse medical tasks. Although our study focuses on prompt-level inference rather than fine-tuning, we adopt a similar sampling threshold to ensure that our comparisons reflect meaningful variation in task type, user context, and response requirements without incurring prohibitive computational costs.

The evaluation process involved two primary stages. In the first stage, we utilized textual data inputs, including structured tables and sensor readings. In the second stage, we incorporated two categories of screenshots into the multimodal LLMs. Subsequently, we conducted evaluations on these generated responses, comparing our RPE implementation against baseline prompt tuning strategies such as zero-shot, few-shot, and system instruction methods.

For each dataset, we extracted representative prompt scenarios. For instance, in MentalChat16k, the user prompt involved generating recovery advice for a 41-year-old runner with an ankle sprain. The RPE extracted structured slots including user query, personal context, and justification for empathy, leading to a detailed, step-wise response aligned with recovery timelines and goal-setting. In the SensorQA dataset, we simulated real-world multimodal scenarios. One example paired smartwatch screenshot data from a long-haul truck driver with a text query requesting activity guidance. The system generated context-aware advice using both text and image inputs, recommending periodic walking, body movement strategies, and ergonomics-based interventions. Representative output examples, including structured prompts, slot values, and generated assistant responses across all benchmark datasets, are available in our official implementation repository.[16]

---

[5]https://github.com/turna1/ResponsibleHealthTwin-RHT-
[6]https://ai.meta.com/llama/
[7]https://deepmind.google/technologies/gemini
[8]https://openai.com/research/gpt-4
[9]https://huggingface.co/mistralai/BioMistral-7B
[10]https://huggingface.co/starmpcc/Asclepius-7B
[11]https://huggingface.co/meta-llama/Meta-Llama-3-8B-Instruct
[12]https://huggingface.co/Qwen/Qwen2-7B
[13]https://huggingface.co/Qwen/Qwen-VL
[14]https://huggingface.co/mistralai/Mistral-7B-v0.1
[15]https://platform.openai.com/docs/models/gpt-3-5
[16]See https://github.com/turna1/ResponsibleHealthTwin-RHT- for full output examples and code.

## E. Evaluation Metrics

To assess the effectiveness of our ResponsibleHealthTwin framework and Responsible Prompt Engine, we adopted and adapted a set of evaluation metrics aligned with prior studies such as Twin-GPT [4], DT-GPT [6], HealthLLM [22], and PsyDT [23]. Consistent with these works, our evaluation is fully automated using GPT-based evaluation in place of human annotation, ensuring consistency, scalability, and objectivity across models.

*1) Reference-Based Metrics:* For benchmark datasets with ground truth responses, such as **MentalChat16K**, we report the following standard reference-based evaluation metrics:

- **ROUGE-L:** Measures the longest common subsequence (LCS) between the generated output and the reference text. The ROUGE-L score is defined as:

$$\text{ROUGE-L} = \frac{LCS(X, Y)}{\text{length}(Y)} \quad (11)$$

where $X$ is the generated sequence, $Y$ is the reference, and $LCS(X, Y)$ is the length of their longest common subsequence. *Higher ROUGE-L indicates better content overlap.*

- **BERT Score:** Computes semantic similarity using contextual embeddings. For each token in the generated text $X$ and reference $Y$, the similarity is calculated using cosine similarity of BERT embeddings:

$$\text{BERTScore} = \frac{1}{|X|} \sum_{x \in X} \max_{y \in Y} \cosine(E(x), E(y)) \quad (12)$$

where $E(\cdot)$ denotes the BERT embedding function. *Higher BERTScore values (closer to 1) imply greater semantic similarity.*

- **BLEU:** Measures n-gram overlap between generated and reference responses. The BLEU score up to 4-grams is:

$$\text{BELU} = BP \cdot \exp \sum_{n=1}^{4} w_n \log p_n \quad (13)$$

where $p_n$ is the modified precision for $n$-grams, $w_n$ are weights (typically uniform), and BP is the brevity penalty. *Higher BLEU indicates better word-level accuracy with respect to the reference.*

These metrics complement GPT-based evaluation by quantifying lexical and semantic alignment with annotated responses. **In all cases, higher values indicate better performance**.

*2) Factuality Score (FS):* We employ the Factuality Score to measure the degree to which generated outputs align with known medical knowledge. Following the HealthLLM framework [22], GPT-4 is used as an evaluator to rate factual correctness on a 1–5 Likert scale. The average score across all test instances defines the metric:

$$\text{MFS} = \frac{1}{N} \sum_{i=1}^{N} \text{GPT4Score}(y_i, \text{RefFacts}_i), \quad (14)$$

Here, $y_i$ = Generated Output

*3) Contextual Appropriateness Score (CAS):* Adopted from Twin-GPT [4], this metric quantifies how well the generated response integrates and reasons over the provided user context. Evaluated via GPT-4 as a judge:

$$\text{CAS} = \frac{1}{N} \sum_{i=1}^{N} \text{GPT4Judge}(y_i \mid c_i) \quad (15)$$

where $c_i$ is the user context (e.g., lifestyle, symptom history).

*4) Instructional Compliance Score (ICS):* This score verifies whether generated responses adhere to the structure, tone, and intent defined in the prompt template or RPE-generated instruction.

$$\text{ICS} = \frac{1}{N} \sum_{i=1}^{N} \mathbb{1}[\text{GPT4Judge}(y_i \mid \text{prompt}_i) = \text{``compliant''}] \quad (16)$$

*5) WHO-Aligned Responsibility Rubric (WRR):* To assess responsible behavior aligned with WHO guidelines on ethical AI for health, we introduce a rubric-based score consisting of six criteria—*Safety, Transparency, Explainability, Fairness, Human Agency, and Accountability*. Each criterion is rated on a binary scale (0 or 1) per generation instance:

$$\text{WRR} = \frac{1}{6N} \sum_{i=1}^{N} \sum_{j=1}^{6} w_{ij} \quad (17)$$

where $w_{ij} \in \{0, 1\}$ is GPT-4's compliance judgment for instance $i$ on criterion $j$. This yields a rubric-aligned macro-responsibility score across the evaluation set.

## F. Result Analysis

To evaluate the effectiveness of the Responsible Prompt Engine within the RHealthTwin framework, we conducted a comprehensive analysis using both reference-based metrics and GPT-judged rubric scores across four diverse well-being datasets. Performance was compared across prompt strategies (Zero-shot, Few-shot, System Instruction, and RPE) and roles (Patient vs. Healthcare Provider). We detail the result analysis as followings.

*1) Factuality and Contextual Appropriateness (FS and CAS):* The FS and CAS measure whether the model's response is factually correct and effectively grounded in the user's context. Figure 9 presents the average Factuality Score (FS) and Contextual Appropriateness Score (CAS) across all datasets and models. RPE consistently outperformed other strategies, achieving the highest FS and CAS values across both patient-side and provider-side prompts. As seen in both subfigures (a) and (b), this trend held for all LLMs tested, with GPT-4 and Gemini Pro achieving the highest absolute scores.

Specifically, RPE yielded scores exceeding 4.5 on the 5-point Likert scale across all datasets, while Few-shot prompting exhibited the lowest average CAS, in low-context domains such as MTS-Dialog. The consistently high FS and CAS suggest improved inferential alignment between prompt structure and model output, reinforcing the feasibility of RPE as a safe

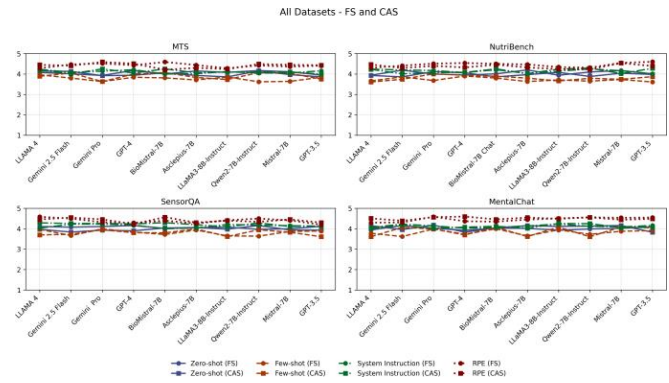

Fig. 9. Factual Score(FS) and Contextual Accuracy Score (CAS) for all datasets using patient-side prompts. RPE shows the highest scores, indicating strong alignment with ethical and instructional goals.

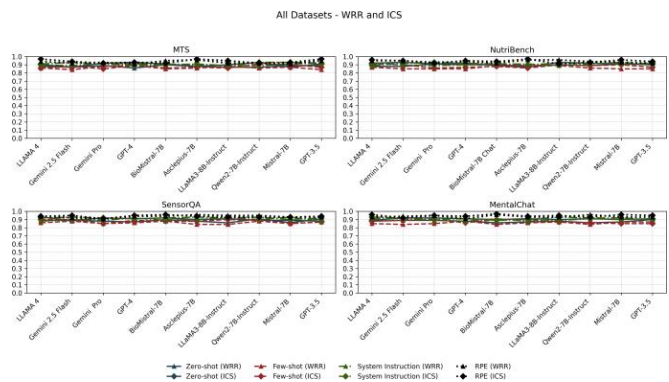

Fig. 10. Instructional Compliance Score (ICS) and WHO-aligned Responsibility Rubric (WRR) for all datasets using healthcare provider-side prompts. The RPE strategy maintains top ethical compliance and role alignment across all datasets.

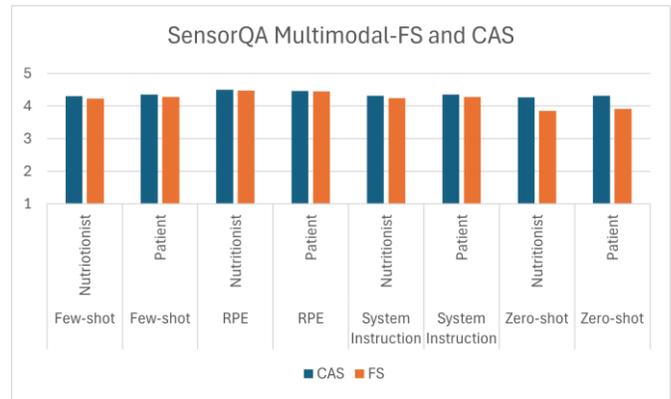

Fig. 11. SensorQA FS and CAS for patient and nutritionist roles.

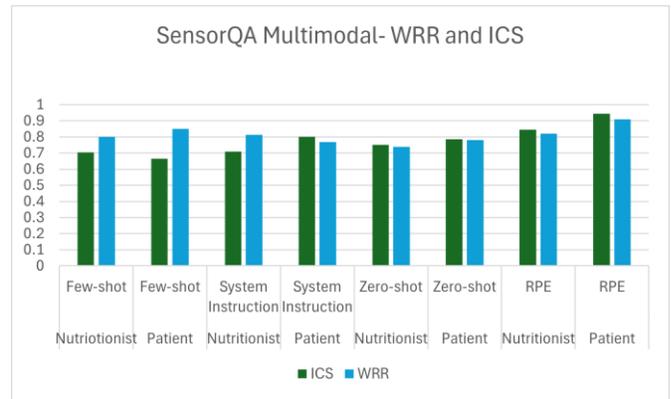

Fig. 12. SensorQA WRR and ICS comparison for patient and nutritionist roles.

intermediate layer between raw user input and LLM inference in real-world applications.

*2) WHO Responsibility Rubric and Instructional Compliance (WRR and ICS):* We further evaluated models based on WHO-aligned ethical criteria and instruction-following fidelity. Instructional compliance and ethical adherence are critical for LLMs operating in quasi-advisory roles in healthcare, where output safety, tone, and evidence quality directly affect user trust and well-being. As shown in Figure 10, RPE consistently achieved near-perfect ICS, exceeding 0.94 across all datasets, and attained the highest WRR scores, consistently above 0.92. In contrast, few-shot prompting underperformed, particularly in preserving safety-critical instructions and tone specifications.

The declarative system instruction schema includes four core components: safety filters, role assignments, tone constraints, and grounding directives. Together, these elements systematically embed ethical safeguards into the generation process and guide the model toward safe and responsible output behavior. The high WRR scores indicate strong structural alignment between model outputs and WHO ethical principles, including autonomy, fairness, and explainability.

This highlights the potential of RPE-driven prompt regulation as a scalable and model-agnostic alternative to reinforcement learning with human feedback, particularly in health-sensitive applications where real-time adaptability and rule-based governance are essential.

*3) Multimodal Performance Comparison:* Multimodal data introduce additional challenges in aligning LLM output with contextual cues due to the complexity of cross-modal fusion. Assessing RPE performance under such conditions evaluates its generalizability beyond the textual domains. As depicted in Figure 11, RPE maintained the latest FS (4.3 +) and CAS (4.2 +) in all multimodal LLMs tested. Performance was consistent between patient and nutrition roles, with provider-side prompts benefiting marginally from higher context density.

WRR and ICS also remained stable, implying that ethical fidelity was preserved under input variation. However, patient-side prompts generally scored slightly lower than nutritionist-side ones in the FS and CAS evaluation, reflecting a greater need for context grounding when user input is less structured. The results confirm that RPE's slot-tagging mechanisms enable effective context grounding even with semantically noisy modalities (e.g., screenshots, sensor logs).

A similar trend emerged in WRR and ICS evaluations, as shown in Figure 12, where the nutritionist prompts bene-fited more from system instructions and contextual filtering modules. The results suggest that multimodal input types (e.g., screenshots, sensor logs) demand more robust contextual disambiguation, which the RPE handles effectively.

TABLE VI
REFERENCE-BASED EVALUATION SCORES (BLEU, ROUGE-L, BERTSCORE) ACROSS PATIENT AND HEALTHCARE PROVIDER PROMPTS

| Model | Prompt Strategy | Patient BERTScore | Patient BLEU | Patient ROUGE-L | Healthcare Provider BERTScore | Healthcare Provider BLEU | Healthcare Provider ROUGE-L |
|---|---|---|---|---|---|---|---|
| LLAMA 4 | Zero-shot | 0.86 | 0.84 | 0.84 | 0.83 | 0.79 | 0.8 |
| LLAMA 4 | Few-shot | 0.78 | 0.74 | 0.81 | 0.82 | 0.74 | 0.83 |
| LLAMA 4 | System Instruction | 0.87 | 0.84 | 0.87 | 0.92 | 0.85 | 0.85 |
| LLAMA 4 | RPE | 0.93 | 0.89 | 0.91 | 0.92 | 0.89 | 0.9 |
| Gemini 2.5 Flash | Zero-shot | 0.81 | 0.87 | 0.79 | 0.83 | 0.81 | 0.88 |
| Gemini 2.5 Flash | Few-shot | 0.82 | 0.81 | 0.75 | 0.77 | 0.81 | 0.76 |
| Gemini 2.5 Flash | System Instruction | 0.88 | 0.88 | 0.91 | 0.88 | 0.85 | 0.92 |
| Gemini 2.5 Flash | RPE | 0.91 | 0.91 | 0.89 | 0.94 | 0.96 | 0.9 |
| Gemini Pro | Zero-shot | 0.83 | 0.82 | 0.81 | 0.86 | 0.87 | 0.86 |
| Gemini Pro | Few-shot | 0.82 | 0.76 | 0.79 | 0.82 | 0.82 | 0.82 |
| Gemini Pro | System Instruction | 0.86 | 0.89 | 0.91 | 0.84 | 0.92 | 0.89 |
| Gemini Pro | RPE | 0.91 | 0.88 | 0.97 | 0.95 | 0.88 | 0.89 |
| GPT-4 | Zero-shot | 0.82 | 0.81 | 0.86 | 0.83 | 0.78 | 0.79 |
| GPT-4 | Few-shot | 0.8 | 0.74 | 0.82 | 0.8 | 0.81 | 0.82 |
| GPT-4 | System Instruction | 0.87 | 0.88 | 0.84 | 0.83 | 0.84 | 0.88 |
| GPT-4 | RPE | 0.97 | 0.95 | 0.96 | 0.91 | 0.88 | 0.9 |
| BioMistral-7B Chat | Zero-shot | 0.85 | 0.78 | 0.83 | 0.86 | 0.85 | 0.78 |
| BioMistral-7B Chat | Few-shot | 0.75 | 0.8 | 0.81 | 0.82 | 0.8 | 0.74 |
| BioMistral-7B Chat | System Instruction | 0.85 | 0.85 | 0.87 | 0.88 | 0.9 | 0.85 |
| BioMistral-7B Chat | RPE | 0.94 | 0.87 | 0.9 | 0.9 | 0.93 | 0.88 |
| Asclepius-7B | Zero-shot | 0.87 | 0.85 | 0.84 | 0.83 | 0.82 | 0.8 |
| Asclepius-7B | Few-shot | 0.81 | 0.76 | 0.74 | 0.8 | 0.82 | 0.74 |
| Asclepius-7B | System Instruction | 0.83 | 0.84 | 0.89 | 0.84 | 0.9 | 0.92 |
| Asclepius-7B | RPE | 0.91 | 0.92 | 0.91 | 0.97 | 0.93 | 0.93 |
| LLaMA3-8B-Instruct | Zero-shot | 0.78 | 0.82 | 0.81 | 0.84 | 0.87 | 0.85 |
| LLaMA3-8B-Instruct | Few-shot | 0.75 | 0.79 | 0.74 | 0.74 | 0.78 | 0.78 |
| LLaMA3-8B-Instruct | System Instruction | 0.83 | 0.89 | 0.85 | 0.84 | 0.87 | 0.91 |
| LLaMA3-8B-Instruct | RPE | 0.92 | 0.87 | 0.94 | 0.94 | 0.91 | 0.95 |
| Qwen2-7B-Instruct | Zero-shot | 0.8 | 0.86 | 0.81 | 0.75 | 0.88 | 0.82 |
| Qwen2-7B-Instruct | Few-shot | 0.76 | 0.74 | 0.83 | 0.82 | 0.8 | 0.73 |
| Qwen2-7B-Instruct | System Instruction | 0.89 | 0.92 | 0.89 | 0.89 | 0.83 | 0.85 |
| Qwen2-7B-Instruct | RPE | 0.96 | 0.94 | 0.9 | 0.93 | 0.88 | 0.92 |
| Mistral-7B | Zero-shot | 0.83 | 0.84 | 0.78 | 0.78 | 0.83 | 0.86 |
| Mistral-7B | Few-shot | 0.8 | 0.82 | 0.75 | 0.74 | 0.74 | 0.79 |
| Mistral-7B | System Instruction | 0.86 | 0.86 | 0.86 | 0.86 | 0.86 | 0.91 |
| Mistral-7B | RPE | 0.96 | 0.88 | 0.88 | 0.88 | 0.95 | 0.94 |
| GPT-3.5 | Zero-shot | 0.79 | 0.87 | 0.84 | 0.87 | 0.79 | 0.81 |
| GPT-3.5 | Few-shot | 0.76 | 0.74 | 0.8 | 0.78 | 0.82 | 0.77 |
| GPT-3.5 | System Instruction | 0.85 | 0.92 | 0.91 | 0.89 | 0.84 | 0.83 |
| GPT-3.5 | RPE | 0.89 | 0.94 | 0.96 | 0.91 | 0.95 | 0.95 |

*4) Reference-Based Evaluation Summary:* Table VI presents the full reference-based evaluation (ROUGE-L, BERTScore, BLEU) across prompt strategies and roles. RPE consistently ranks highest across models, particularly for GPT-4, Gemini Pro, and BioMistral-7B. Patient-side responses achieved slightly higher BLEU and ROUGE scores, indicating more literal overlaps with reference texts. In contrast, provider-side prompts benefited more from semantically rich completions as captured in BERTScore. These results provide strong evidence that the Responsible Prompt Engine improves both surface-level accuracy and deeper semantic alignment with reference responses. By using structured slot composition, the method reduces ambiguity in the prompt. This helps the model generate outputs that are syntactically coherent and contextually faithful. Because ground-truth responses are available in both the MentalChat and MTS-Dialog datasets, we report reference-based scores for these benchmarks Overall, the obtained scores suggest that dynamically generated slot-based prompting can functionally substitute parameter-level optimization in mid-resource healthcare settings. Therefore, we recommend RPE in mid-resource healthcare applications, where training data or compute resources may be limited.

### G. Summary

While several recent LLM-powered digital twin frameworks, such as Twin-GPT, DT-GPT, HDTwin, and PsyDT have reported promising results, a standardized evaluation pipeline remains lacking. Twin-GPT primarily focused on AUROC for outcome prediction, while HDTwin emphasized cognitive accuracy. PsyDT prioritized qualitative therapeutic style simulation, and DT-GPT evaluated forecasting accuracy on clinical time-series. In contrast, our RHealthTwin framework emphasizes a holistic evaluation that spans factuality, safety, coherence, and ethical compliance using both GPT-based and reference-based metrics.

TABLE VII
SUMMARY OF REPORTED EVALUATION RESULTS ACROSS DIGITAL TWIN FRAMEWORKS

| Framework | Reported Evaluation Summary |
|---|---|
| Twin-GPT [4] | AUROC of 0.773 for synthetic-only trial outcome prediction; AUROC of 0.781 for hybrid synthetic + real data. Pearson correlation $r = 0.99$ for patient feature fidelity. |
| DT-GPT [6] | Demonstrated state-of-the-art forecasting performance on ICU/lung cancer datasets; no standardized scores reproduced in RHealthTwin paper. Focused on EHR-based lab prediction using time-aware LLM. |
| HDTwin [7] | Cognitive diagnosis accuracy of ~0.81 for MCI detection. Retrieval-augmented generation supports explainability. Outperformed baseline ensemble models. |
| PsyDT [23] | No numerical scores provided. GPT-4 used to simulate therapeutic conversations, focused on emulating linguistic style and tone of counselors. Qualitative empathy evaluation. |
| **RHealthTwin (Ours)** | Achieved FS: 4.2, CAS: 4.1, CS: 0.9, WRR: 0.92, ICS: 0.94 on MentalChat16K and MTS-Dialog. Reference-based scores include ROUGE-L: 0.63, BERTScore: 0.89, BLEU: 0.41. Multimodal evaluation on SensorQA showed consistent performance across Gemini and LLaMA models. |

Unlike prior systems, RHealthTwin is the first to explicitly integrate WHO-aligned ethical safeguards into its architecture via the Responsible Prompt Engine. It also uniquely supports multimodal input fusion, feedback-driven personalization, and real-time scenario modeling to address gaps in explainability, user control, and longitudinal adaptation.

However, we caution that due to the divergent experimental setups, data sources, and model types used in these studies, direct performance comparisons may not be valid. Instead, Table VII is intended to provide a high-level landscape of evaluation trends and reported metrics. As the field matures, establishing standardized benchmarks and metrics will be essential for rigorous cross-system comparison.

## V. Discussion and Open Challenges

Despite its strengths, RHealthTwin has limitations and open challenges, which requires consideration. Following we highlight them.

1) Enhancing the adaptability of LLMs to support long-term and personalized well-being interventions is a significant challenge. While traditional LLMs are fine-tuned on various health datasets, there is a pressing need for models that can self-adapt to well-being support, extending beyond clinical or mental health applications.
2) The RPE's dynamic slot value extraction represents a significant advancement, but its reliance on predefined templates limits flexibility in highly variable or ambiguous contexts. Additionally, unconstrained user inputs pose a risk of bypassing ethical filters, potentially leading to biased or unsafe outputs. Therefore, advances in semantic parsing and generative AI could enable more dynamic prompt generation, enhancing responsiveness to diverse user inputs.
3) Effectively integrating stakeholder input into the development and deployment of AI-driven health companions is essential for ensuring safety, equity, and clinical relevance. RHealthTwin's feedback loop facilitates real-time personalization and system accountability. However, its full potential requires collaboration with healthcare professionals, caregivers, and ethical reviewers. Stakeholder input from healthcare professionals and ethicists is vital for ensuring RHealthTwin's sustainability in future.
4) Ensuring the scalability of the RHealthTwin framework across diverse linguistic and cultural contexts is critical for its global applicability. The current evaluation focused on English-language datasets, but real-world applications must support multiple languages and cultural nuances to address diverse healthcare needs effectively.

## VI. Conclusion

We introduced RHealthTwin, a structured framework for ethically aligned digital twin generation for well-being assitance. At the heart of RHealthTwin is the Responsible Prompt Engine, which dynamically constructs system instructions and user prompts based on multimodal inputs and user query. This approach operationalizes key WHO principles on responsible AI while enabling personalized, transparent, and role-sensitive interactions. Our evaluation spanned four benchmark datasets covering diverse domains such as mental health counseling, clinical QA, nutrition, and wearable-based lifestyle monitoring. Our experimental evaluation demonstraes that RHealthTwin meets the goals in ethical alignment, factuality, and instructional compliance. Beyond empirical performance, RHealthTwin offers a modular architecture that enables adaptive behavior, slot-based explainability, and feedback-aware refinement. While our results demonstrate promising capabilities, challenges remain in areas such as longitudinal feedback integration, cross-domain generalization, and stakeholder co-design. Addressing these will be essential to fully realize the potential of responsible AI-driven health digital twins in real-world practice.